\documentclass[11pt,a4paper]{article}
\usepackage{authblk}
\usepackage[utf8]{inputenc}
\usepackage[T1]{fontenc}
\usepackage[english]{babel}
\usepackage{lmodern}
\usepackage{geometry}
\usepackage{microtype}
\usepackage{booktabs}
\usepackage{tabularx}
\usepackage{longtable}
\usepackage{array}
\usepackage{ragged2e}
\usepackage{enumitem}
\usepackage{amsmath,amssymb}
\usepackage{tikz}
\usetikzlibrary{arrows.meta,positioning,shapes.geometric}
\usepackage{hyperref}
\usepackage{xcolor}
\usepackage{caption}
\usepackage{graphicx}
\usepackage{placeins}
\usepackage{float}

\renewcommand{\arraystretch}{1.18}
\hypersetup{colorlinks=true,linkcolor=black,citecolor=black,urlcolor=blue}
\newcolumntype{Y}{>{\RaggedRight\arraybackslash}X}
\newcolumntype{P}[1]{>{\RaggedRight\arraybackslash}p{#1}}

\title{\textbf{Towards a Unified Taxonomy of Agricultural Harvesting Technologies: Linking Physical Interaction, Automation, and Selectivity}}
\author[1]{Renato Cristiano Torres}
\date{October 5, 2026}
\author[2]{Fabiano Mariath D'Oliveira}
\author[3]{Claudio Gottschalg Duque }
\date{October 5, 2026}

\affil[1]{Embrapa Agricultura Digital, Research Group on Scientific Computing, Information Engineering, and Automation, Campinas, Brazil}
\affil[2]{Embrapa, Information Technology Departament,  Federal District, Brazill}
\affil[3]{University of Brasilia,  Computational Linguistics, Federal District, Brazil}
\begin{document}
\maketitle

\begin{abstract}
Harvesting technologies have evolved from conventional mechanized systems based on bulk mechanical interaction toward increasingly automated and robotic systems capable of perception, decision-making, and localized actuation. However, this technological evolution has also created a classification problem. Harvesting systems are commonly described according to crop type, mechanical removal principle, sensing technology, robotic architecture, or degree of automation, making direct comparison between mechanized, automated, and robotic approaches difficult.

This paper argues that a unified taxonomy is needed to describe agricultural harvesting technologies across different levels of technological complexity. Two primary classification axes are identified as fundamental: the physical principle governing plant--machine interaction and the level of automation incorporated into the harvesting process. Selectivity is treated as an emergent functional attribute that indicates the biological and spatial scale at which perception and decision-making influence physical actuation.

Based on these classification axes and the selectivity attribute, conventional mechanized methods including cutting, pulling, vibration, impact, brushing, pneumatic action, and combined systems are analyzed together with semi-automated and robotic harvesting approaches. The analysis shows that increasing automation does not necessarily result in greater harvesting selectivity. A qualitative transition occurs when perception and decision-making directly influence product-level actuation, shifting the harvesting process from distributed or bulk interaction toward localized and selective manipulation.

The proposed framework provides a common conceptual basis for positioning and comparing mechanized, automated, and robotic harvesting technologies. By separating the physical mechanism of product removal from the intelligence and selectivity of the harvesting system, the taxonomy helps clarify technological boundaries, identify intermediate levels of automation, and support the analysis and development of future agricultural harvesting systems.
\end{abstract}

\section{Introduction}

The adoption of agricultural mechanization technologies must be understood within a broader historical, economic, and social context. In a comparative analysis of the history of agricultural mechanization, Binswanger \cite{ref5} demonstrates that the introduction of machinery in agricultural systems does not occur in a linear manner, nor is it driven exclusively by technological availability. Factors such as relative labor costs, land tenure structure, capital availability, public policies, and agronomic characteristics of crops play a decisive role in the adoption or rejection of mechanized and automated technologies. These factors shape not only the feasibility of technological implementation but also its economic and operational sustainability across different production systems.

Mechanization is defined as the use of mechanical energy and machinery to replace or reduce human or animal labor \cite{ref12}, whereas automation refers to systems capable of monitoring, decision-making, and actuation through sensors, controllers, and actuators \cite{ref14}. Although conceptually distinct, these two paradigms increasingly coexist in modern agricultural systems.

Harvesting is among the most labor-intensive and costly operations in agriculture, particularly in high-value horticultural and fruit production systems. Previous studies indicate that harvesting may account for approximately 30--50\% of total production costs \cite{ref3,ref4}. Labor shortages and rising wages have intensified the adoption of agricultural automation technologies, a trend further reinforced by changes in consumer behavior, which increasingly demand consistent and high-quality products \cite{ref17}.

Unlike operations such as seeding and spraying, harvesting involves direct interaction with fragile biological materials, requiring simultaneous perception, decision-making, and precise mechanical manipulation \cite{ref25}. The heterogeneity of harvested products in terms of size, shape, rigidity, and maturity stage significantly increases the complexity of harvesting operations. Despite advances in robotics, computer vision, and artificial intelligence, economic, operational, and crop-adaptation challenges continue to limit the widespread diffusion of automated harvesting technologies.

Modern agricultural machines increasingly incorporate GPS-based navigation, electronic controllers, onboard sensors, and yield monitoring systems. However, the incorporation of these technologies does not automatically change the physical principle used to harvest the crop. A machine may become more automated while still applying the same cutting, pulling, vibration, impact, or combined harvesting mechanism. In this sense, mechanization and automation are not mutually exclusive categories but different dimensions of a harvesting system.

This distinction creates a conceptual problem for comparison. Conventional harvesting machinery is usually described primarily by the physical mechanism used to detach, collect, or transport the product, whereas robotic harvesting systems are more frequently discussed in terms of perception, manipulation, motion planning, and autonomy. As a consequence, technologies belonging to different stages of harvesting automation are often described using classification criteria that are not directly comparable.

The problem becomes even more evident because increasing automation does not necessarily imply increasing selectivity. Automated navigation or automatic adjustment of process parameters may improve operational efficiency while preserving bulk harvesting. A qualitative transition occurs only when perception and decision-making begin to influence product-level actuation, allowing the system to discriminate between harvestable and non-harvestable targets.

Accordingly, this study argues that agricultural harvesting technologies require a unified classification framework capable of describing both their physical interaction mechanisms and their level of automation while explicitly evaluating selectivity. Two primary classification axes are adopted: (i) the physical principle governing plant--machine interaction during product removal and (ii) the degree of automation associated with perception, decision-making, and actuation. Selectivity is treated as an emergent functional attribute that expresses the biological and spatial scale at which the harvesting action is targeted.

The objective of this paper is therefore not only to review existing harvesting technologies, but to demonstrate the conceptual need for a unified taxonomy capable of positioning mechanized, automated, and robotic harvesting systems within a common analytical framework. The main contributions are:
\begin{enumerate}[label=(\roman*)]
    \item to separate the physical mechanism of product removal from the automation level of the harvesting system;
    \item to identify selectivity as a central linking attribute between physical interaction and automation;
    \item to organize conventional mechanized, semi-automated, collaborative, and autonomous robotic harvesting within a common framework; and
    \item to clarify the technological transition from distributed or bulk harvesting toward localized and selective product-level interaction.
\end{enumerate}

\section{The Classification Problem in Agricultural Harvesting}

Agricultural harvesting technologies span a wide range of mechanical and computational complexity. At one end are conventional machines designed to remove, separate, or collect products through mechanical forces applied to the crop as a whole or to large portions of the plant. At the other end are robotic systems that identify individual products, estimate harvest readiness and pose, plan motions, and execute localized cutting, grasping, or suction actions \cite{ref19,ref21,ref24,ref27,ref35,ref44}.

These two technological domains are often discussed using different vocabularies. Agricultural machinery literature typically emphasizes mechanical principles, field efficiency, crop architecture, machine stability, and compatibility with production geometry \cite{ref9,ref22,ref23,ref39}. Robotic harvesting literature, in contrast, emphasizes sensing, visual perception, target recognition, manipulation, planning, control, and autonomy \cite{ref19,ref21,ref24,ref25,ref26,ref35,ref44}. Both descriptions are valid, but neither dimension alone is sufficient to position the full range of contemporary harvesting technologies.

A physical classification alone cannot distinguish a manually controlled vibrating harvester from a sensor-regulated vibration system. Conversely, an automation-based classification alone does not reveal whether the product is detached by cutting, pulling, vibration, impact, grasping, or suction. Moreover, systems with similar automation levels may differ significantly in selectivity. This produces overlapping categories and makes direct comparison difficult.

The need for a unified taxonomy therefore arises from three observations. First, the physical harvesting principle and the level of automation are conceptually distinct. Second, mechanization can incorporate automation without becoming selective. Third, selectivity depends on whether perception and decision-making affect the location and target of physical actuation. A useful taxonomy must therefore preserve these distinctions rather than collapse them into a single linear scale.

Table~\ref{tab:priorreviews} illustrates the fragmentation of emphasis across representative reviews cited in this manuscript. The comparison is not intended as a ranking of previous studies, but as a conceptual synthesis of the principal dimensions emphasized in their reported scopes. Mechanization-oriented reviews tend to foreground harvesting mechanisms and crop--machine compatibility, whereas robotic-harvesting reviews tend to foreground perception, manipulation, planning, and autonomy. Selectivity is explicit in some robotic reviews, but is not generally integrated with physical harvesting principles and automation level within a single classification structure.

\begin{table}[!htbp]
\centering
\caption{Primary emphasis of representative harvesting reviews and the integrative scope of the present framework. ``Partial'' indicates that the dimension is discussed but is not the principal organizing axis of the review.}
\label{tab:priorreviews}
\small
\resizebox{\textwidth}{!}{%
\begin{tabular}{lccccc}
\toprule
\textbf{Reference} & \textbf{Physical interaction} & \textbf{Perception/robotics} & \textbf{Automation} & \textbf{Selectivity} & \textbf{Unified cross-domain taxonomy} \\
\midrule
Droukas et al. \cite{ref7} & Partial & Strong & Strong & Partial & -- \\
Kootstra et al. \cite{ref24} & Partial & Strong & Strong & Strong & -- \\
Pu \cite{ref34} & Strong & Partial & Partial & Partial & -- \\
Rajendran et al. \cite{ref35} & Partial & Strong & Strong & Strong & -- \\
Zhang \cite{ref44} & Partial & Strong & Strong & Strong & -- \\
Present framework & Strong & Strong & Strong & Explicit attribute & \checkmark \\
\bottomrule
\end{tabular}%
}
\end{table}

\section{Review Scope and Analytical Approach}

This study is a conceptual comparative review rather than a systematic review. It synthesizes agricultural engineering, mechanization, sensing, and harvesting robotics literature represented in the cited references to identify recurring physical and functional principles used to describe harvesting systems. The manuscript does not claim exhaustive coverage of all harvesting machines or robotic platforms.

The analysis follows three steps. First, the literature is examined to identify the main physical principles used for product removal and the principal mechanical and biological constraints associated with their application. Second, automated and robotic harvesting approaches are examined in terms of perception, decision-making, and actuation, with particular attention to the emergence of localized product-level interaction. Third, these observations are combined into a unified conceptual framework structured around two primary axes---physical interaction and automation level---with selectivity evaluated as an emergent functional attribute.

This approach is intended to support conceptual comparison across technology classes rather than to rank individual commercial systems or estimate performance statistically. Consequently, the proposed taxonomy should be interpreted as an analytical framework that can be refined as new harvesting mechanisms and robotic architectures emerge.

\section{Mechanical and Biological Constraints of Harvesting}

Mechanized harvesting requires production systems to meet a set of technical conditions that range from soil characteristics to planting layout, directly influencing the feasibility of implementation. These requirements reflect the strong coupling between machine operation and field geometry, as well as the mechanical interaction between harvesting systems and plant structures.

Table~\ref{tab:requirements} summarizes representative technical ranges and requirements reported in the agricultural engineering literature. These values should not be interpreted as universal design limits because their applicability depends on crop, machine architecture, terrain, and production system.

\begin{table}[!htbp]
\centering
\caption{Representative production-system requirements reported for mechanized harvesting.}
\label{tab:requirements}
\small
\begin{tabularx}{\textwidth}{P{0.22\textwidth} Y P{0.20\textwidth}}
\toprule
\textbf{Requirement} & \textbf{Representative range or condition reported in the literature} & \textbf{References} \\
\midrule
Terrain slope & Lower than approximately 12--15\%, supporting machine stability and operational safety & \cite{ref1,ref8,ref22,ref29,ref30,ref32,ref43} \\
Row spacing & Crop-dependent spacing allowing adequate machine traffic and harvesting operation & \cite{ref4,ref9,ref22} \\
Plant spacing & Approximately 0.5--0.7 m in reported applications requiring compatibility between plant size and mechanized systems; not a general limitation for grain and seed harvesting & \cite{ref12,ref22,ref39} \\
Obstacles between rows & Removal or management of stones, stumps, crop residues, and other obstacles that may interfere with machine displacement & \cite{ref13,ref22,ref23,ref39,ref41} \\
Planting row length & Long rows are generally advantageous because they reduce turning maneuvers and improve operational efficiency & \cite{ref22,ref23,ref39} \\
Headland width & Approximately 6.0 m in reported cases to allow machine maneuvering & \cite{ref9,ref22,ref23,ref39} \\
Plant height & Training systems with plant height around or below 3.0 m in reported fruit-harvesting applications & \cite{ref8,ref9,ref22,ref39} \\
Minimum branch clearance & Approximately 0.40 m from the soil in reported mechanized systems to reduce mechanical interference & \cite{ref8,ref9,ref22,ref39} \\
\bottomrule
\end{tabularx}
\end{table}

The constraints in Table~\ref{tab:requirements} indicate that mechanized harvesting does not depend solely on technological availability, but also on careful planning and physical configuration of the production system. They also illustrate why the physical-interaction axis of the taxonomy cannot be interpreted independently of crop architecture and production-system structure. Horticultural and fruit crops, in particular, exhibit morphological and architectural characteristics that often require combined use of mechanization and automation to ensure effective harvesting while preserving product quality and plant integrity.

Fruit harvesting commonly involves separation at the peduncle--fruit interface, whose mechanical resistance depends on species, maturity stage, and environmental conditions. Detachment resistance varies substantially with crop, maturity stage, plant architecture, and the harvesting mechanism used \cite{ref2,ref6,ref34}. Fruit tissues are also susceptible to mechanical damage under bending, compression, and impact loading \cite{ref18}.

The physiological state of the plant directly affects both detachment force and fruit fragility. Optical and spectral signals have been widely investigated as non-destructive indicators of fruit maturity and quality \cite{ref31,ref46}. These signals are therefore relevant for sensing strategies and for determining appropriate manipulation actions during harvesting.

Overall, the literature indicates that conventional mechanized harvesting generally relies more strongly on crop and field standardization, whereas advanced automation increasingly transfers part of the adaptation burden from the production system to the harvesting machine. This distinction is gradual rather than absolute: automated systems may still require structured production environments, while mechanized systems can incorporate adaptive control.

\section{From Bulk Interaction to Selective Harvesting}

From the perspective of robotics applied to harvesting, selectivity is not merely a desirable feature but a fundamental property of the interaction between the harvesting system and biological conditions. In traditional mechanized harvesting methods, such as cutting, pulling, and vibration, product removal is generally achieved through distributed mechanical forces applied to portions of the plant without an inherent capability to discriminate among individual plant organs. This interaction paradigm results in low selectivity and strong dependence on crop standardization.

In contrast, harvesting systems that incorporate advanced automation and robotics employ perception and control mechanisms that enable localized interactions with the plant based on individual identification of the target product. Recent reviews emphasize that computer-vision-based perception is central to selective harvesting because it supports assessment of harvest readiness and decisions regarding when and where harvesting actions should be performed \cite{ref21,ref24,ref35}. Selective Harvesting Robots integrate sensing, motion planning, and mechanical actuation to operate adaptively in heterogeneous environments \cite{ref35}.

Research on selective harvesting robotics highlights strategies including fruit maturity detection, estimation of position and orientation for cutting or grasping, machine-learning-based recognition, and active perception under occlusion or changing viewpoints \cite{ref7,ref28,ref35}. These capabilities do not merely automate movement; they alter the spatial scale at which the harvesting action is decided.

This distinction explains why incremental automation in mechanized harvesting systems---such as assisted navigation or automatic control of height, speed, or vibration---does not necessarily alter harvesting selectivity. A qualitative shift occurs when automation directly influences product identification and decision-making and is coupled to localized actuation. Thus,
\begin{equation}
\text{high automation} \;\not\Rightarrow\; \text{high selectivity}.
\end{equation}

By contrast, the potential for selective harvesting emerges when perception and decision-making influence the target and location of actuation:
\begin{equation}
\text{perception} + \text{decision} + \text{localized actuation}
\;\longrightarrow\; \text{potential for selective harvesting}.
\end{equation}
\begin{figure}[h]
    \centering
    \includegraphics[width=1\linewidth]{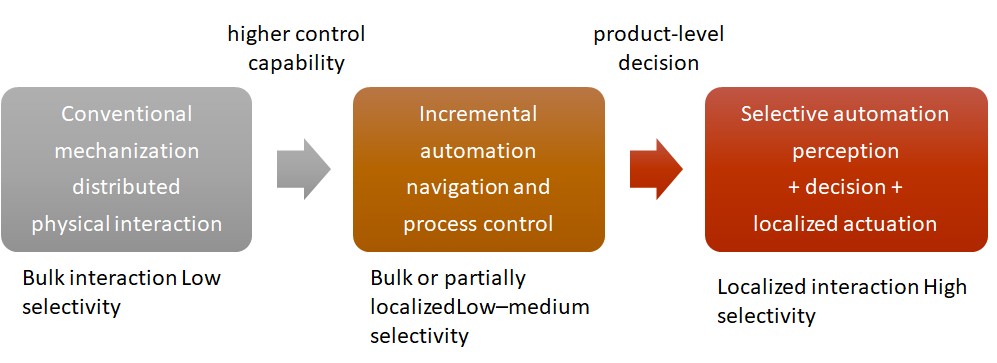}
    \caption{Conceptual transition from conventional bulk harvesting to selective product-level
harvesting. Increasing automation alone does not guarantee selectivity; the decisive transition
occurs when perception and decision-making govern localized actuation. Source: authors’
conceptual synthesis.}
    \label{fig:placeholder}
\end{figure}

\section{Dimensions Required for a Unified Harvesting Taxonomy}

The classification problem can be addressed by treating harvesting technology as a multidimensional concept rather than as a single mechanization-to-robotics continuum. In the proposed framework, a harvesting system is positioned by two primary coordinates,
\begin{equation}
\mathbf{H}=(P,A),
\end{equation}
where $P$ denotes the physical interaction principle and $A$ the automation level. Selectivity, $S$, is evaluated as an emergent functional attribute rather than as an independent axis. Conceptually,
\begin{equation}
S = g(P,A,C,E),
\end{equation}
where $C$ represents crop and biological characteristics and $E$ represents environmental and operating conditions. The notation is conceptual rather than a fitted mathematical model; it emphasizes that achieved selectivity depends on more than automation level alone.

The first dimension describes how mechanical energy is transferred to the crop or product during harvesting. The principal mechanisms identified in the reviewed literature include cutting, pulling, vibration, impact, brushing, pneumatic action, and combinations of several operations. Robotic systems may additionally employ localized grasping, selective cutting, and assisted suction.

The second dimension describes which parts of the harvesting process are performed automatically. Automation may be limited to navigation, extend to automatic adjustment of process variables, support partial product selection, or integrate perception, decision-making, planning, and actuation in autonomous harvesting. Human--robot collaborative systems constitute a distinct configuration because task allocation is shared dynamically between human operators and automated subsystems.

Selectivity describes the extent to which the system can discriminate among harvestable and non-harvestable targets and direct actuation toward the intended biological unit. It is not assumed to increase automatically with automation. Instead, it emerges from the interaction among physical mechanism, perception, decision-making, crop characteristics, and operating conditions.

For qualitative comparison in this review, selectivity is interpreted as follows: \textit{low} indicates no product-level discrimination; \textit{medium} indicates partial or conditional target discrimination, often with human intervention or limited automated selection; \textit{high} indicates automated product-level discrimination coupled to target-specific actuation; and \textit{variable} denotes configurations whose selectivity depends strongly on task allocation, crop, or operating mode. These categories are analytical descriptors rather than standardized quantitative thresholds. Taken together, the two primary axes and the selectivity attribute allow systems using the same mechanical principle to be distinguished by automation and achieved selectivity, and systems with similar automation to be distinguished by their physical interaction mechanisms.

\section{Proposed Unified Taxonomy}

The proposed taxonomy integrates the two primary classification axes identified in the previous section: physical interaction and automation level, while considering selectivity as an emergent functional attribute. This structure allows mechanized, automated, and robotic harvesting systems to be positioned within a common framework, even when they differ in mechanical principle, sensing capability, decision-making, or actuation strategy. The purpose is not to replace existing classifications, but to provide a unified reference for comparing technologies that are frequently described using different technical criteria, summarized in the following tables (3,4 e 5).

\begin{table}[!htbp]
\centering
\caption{Mechanized harvesting methods according to physical interaction principle and typical crops.}
\label{tab:mechanized}
\small
\begin{tabularx}{\textwidth}{P{0.16\textwidth} Y P{0.27\textwidth} P{0.14\textwidth}}
\toprule
\textbf{Method} & \textbf{Physical principle} & \textbf{Typical crops} & \textbf{References} \\
\midrule
Cutting & Product separation by mechanical shearing of the stem or entire plant & Grains and seeds (wheat, rice), sugarcane, forage crops & \cite{ref22,ref37,ref39} \\
Pulling & Product removal by tensile force between plant and soil & Roots and tubers (potato, carrot, beetroot) & \cite{ref39} \\
Vibration & Mechanical excitation used to detach the product from the plant & Woody crops (coffee), fruit crops (citrus, olive) & \cite{ref8,ref10,ref34} \\
Impact & Transfer of kinetic energy through repeated impacts & Woody and industrial fruit crops (coffee, olive) & \cite{ref10,ref22,ref34} \\
Brushing & Continuous mechanical contact using flexible elements & Fruit and woody crops under specific production systems & \cite{ref8,ref10,ref15} \\
Pneumatic & Pressure differentials and airflow used for capture or transport & Seeds, lightweight fruits, auxiliary harvesting systems & \cite{ref22,ref34,ref39} \\
Combined & Integration of cutting, threshing, separation, and cleaning in a single system & Large-scale grain and seed crops & \cite{ref22,ref37,ref39} \\
\bottomrule
\end{tabularx}
\end{table}

Mechanized harvesting methods are therefore primarily distinguishable through the physical principle used to remove the product. Their performance depends strongly on crop geometry, product tolerance to mechanical stress, and compatibility between plant architecture and harvesting mechanism.

\begin{table}[!htbp]
\centering
\caption{Automated and robotic harvesting approaches according to physical interaction principle.}
\label{tab:automated}
\small
\begin{tabularx}{\textwidth}{P{0.20\textwidth} Y P{0.27\textwidth} P{0.14\textwidth}}
\toprule
\textbf{Method} & \textbf{Physical principle and automation function} & \textbf{Typical crops} & \textbf{References} \\
\midrule
Selective grasping & Controlled mechanical force applied by robotic manipulators based on perception of the target product & Apple, orange, strawberry, tomato, bell pepper & \cite{ref4,ref19,ref44} \\
Selective cutting & Localized cutting of the peduncle using automated blades or actuators & Lettuce, broccoli, high-value fruit crops & \cite{ref34,ref44} \\
Assisted suction & Pressure differentials used for capture and transport, combined with visual perception & Strawberry, seeds, seedlings, light products & \cite{ref34,ref44} \\
Semi-automated vibration & Mechanical excitation whose parameters are adjusted automatically using sensors and control algorithms & Citrus, olive, woody crops & \cite{ref8,ref34} \\
Human--robot collaboration & Physical and cognitive interaction with task sharing between human operators and automated systems & Horticultural and fruit crops & \cite{ref4,ref15,ref36} \\
Automated harvesting platforms & Automated navigation and operator assistance integrated with harvesting operations & Horticultural and fruit crops & \cite{ref4,ref20,ref34,ref45} \\
\bottomrule
\end{tabularx}
\end{table}

\begin{table}[H]
\centering
\caption{Automation classes in agricultural harvesting and their typical selectivity.}
\label{tab:automation}
\footnotesize
\setlength{\tabcolsep}{4pt}
\renewcommand{\arraystretch}{1.05}
\begin{tabularx}{\textwidth}{P{0.22\textwidth} Y P{0.16\textwidth} P{0.16\textwidth}}
\toprule
\textbf{Class} & \textbf{Description} & \textbf{Typical selectivity} & \textbf{References} \\
\midrule
Mechanization (no automation) & Harvesting execution based on mechanical energy; perception, decision-making, and control performed by the operator & Low & \cite{ref12,ref22,ref39} \\
Automated navigation & Automation limited mainly to machine guidance and displacement while the underlying harvesting method remains mechanized & Low & \cite{ref9,ref22,ref33} \\
Automated control & Automatic adjustment of harvesting parameters such as height, speed, or vibration without individual product selection & Low to medium & \cite{ref9,ref11,ref39} \\
Semi-automated harvesting & Partial product selection based on automated perception with human intervention in decision-making or execution & Medium & \cite{ref4,ref36,ref34} \\
Autonomous robotic harvesting & Integration of perception, decision-making, planning, and actuation in an automated harvesting system & High & \cite{ref4,ref16,ref26,ref38,ref42,ref44} \\
Human--robot collaborative harvesting & Dynamic task sharing between human operators and automated or robotic systems & Variable & \cite{ref4,ref15,ref36} \\
\bottomrule
\end{tabularx}
\end{table}
\renewcommand{\arraystretch}{1.18}

The central implication of the proposed taxonomy is that physical mechanism and automation level should be treated as independent axes, while selectivity is evaluated as an additional functional attribute. Table~\ref{tab:matrix} illustrates how similar physical principles may occur at different automation levels.

\begin{table}[H]
\centering
\scriptsize
\setlength{\tabcolsep}{3pt}
\renewcommand{\arraystretch}{1.0}
\caption{Conceptual matrix combining physical interaction and automation level. Check marks indicate combinations represented or directly supported by the harvesting approaches discussed in the reviewed literature; a dash indicates that the combination is not established by the present review.}
\label{tab:matrix}
\small
\resizebox{\textwidth}{!}{%
\begin{tabular}{lccccc l}
\toprule
\textbf{Physical interaction} & \textbf{Mechanized} & \textbf{Auto. nav./control} & \textbf{Semi-auto.} & \textbf{Autonomous} & \textbf{Typical selectivity} & \textbf{Representative refs.} \\
\midrule
Cutting & $\checkmark$ & $\checkmark$ & $\checkmark$ & $\checkmark$ & Low--high & \cite{ref22,ref34,ref44} \\
Pulling & $\checkmark$ & $\checkmark$ & -- & -- & Low--medium & \cite{ref39} \\
Vibration & $\checkmark$ & $\checkmark$ & $\checkmark$ & -- & Low--medium & \cite{ref8,ref10,ref34} \\
Impact & $\checkmark$ & $\checkmark$ & -- & -- & Low & \cite{ref10,ref22,ref34} \\
Brushing & $\checkmark$ & $\checkmark$ & -- & -- & Low--medium & \cite{ref8,ref10,ref15} \\
Pneumatic/suction & $\checkmark$ & $\checkmark$ & $\checkmark$ & $\checkmark$ & Low--high & \cite{ref22,ref34,ref44} \\
Grasping & -- & -- & $\checkmark$ & $\checkmark$ & Medium--high & \cite{ref4,ref19,ref38,ref42,ref44} \\
Combined systems & $\checkmark$ & $\checkmark$ & $\checkmark$ & $\checkmark$ & Variable & \cite{ref22,ref37,ref39} \\
\bottomrule
\end{tabular}%
}
\end{table}
\renewcommand{\arraystretch}{1.18}

The matrix should be interpreted as a conceptual synthesis rather than as an exhaustive inventory of commercial machines. Its purpose is to show why a single classification dimension is insufficient: a physical harvesting principle may remain unchanged while sensing, decision-making, control, and selectivity evolve substantially.

\section{Implications for Agricultural Robotics and Harvesting-System Design}

The proposed framework has several implications for the analysis and development of harvesting technologies. First, it avoids treating mechanization and robotics as mutually exclusive categories. A mechanized platform may incorporate automated navigation and process control while retaining a conventional physical harvesting principle. Likewise, a robotic platform may still depend on structured crop geometry and may not necessarily achieve high selectivity under all operating conditions.

Second, the framework clarifies that autonomy and selectivity should not be used as synonyms. Navigation autonomy can increase without changing the biological target of the harvesting action. High selectivity generally requires the perception--decision--actuation chain to reach the product level.

Third, the taxonomy provides a common language for comparing systems that would otherwise be described using different technical criteria. For example, a vibration-based system can be classified simultaneously by its physical mechanism, its automation level, and its selectivity. The same logic can be applied to cutting, suction, grasping, and combined systems.

Fourth, the framework can support technology-development decisions. When a harvesting system exhibits poor performance, the classification helps distinguish whether the limitation originates primarily from the physical interaction mechanism, insufficient perception, limited decision-making capability, inadequate actuation, or a mismatch between desired and achieved selectivity.

Finally, the taxonomy highlights intermediate technological configurations that may be overlooked by a simple mechanized-versus-robotic dichotomy. Semi-automated and human--robot collaborative systems may represent practical transitions between conventional machinery and fully autonomous selective harvesting, particularly in crops where complete autonomy remains technically or economically difficult \cite{ref4,ref15,ref36}.

\section{Conclusion}

Agricultural harvesting presents a classification challenge because contemporary systems combine physical mechanisms, sensing, control, decision-making, and robotic actuation at different levels. Conventional mechanized systems are commonly described according to how mechanical energy is applied to the crop, whereas automated and robotic systems are frequently described according to perception, planning, manipulation, and autonomy. These perspectives are complementary but are difficult to compare when used independently.

This study therefore reframed the harvesting problem around the need for a unified taxonomy. The proposed framework distinguishes two primary classification axes---physical interaction principle and automation level---while treating selectivity as an emergent functional attribute. This separation makes it possible to recognize that a harvesting mechanism can remain physically unchanged while its control architecture becomes increasingly automated, and that increasing automation does not necessarily imply selective harvesting.

The transition from bulk harvesting to selective harvesting is better characterized by the integration of perception and decision-making with localized product-level actuation. In this sense, selectivity serves as a functional bridge between the mechanical and computational dimensions of a harvesting system.

The framework organizes conventional mechanized, automated, semi-automated, collaborative, and autonomous robotic approaches within a common analytical space. Its principal contribution is therefore not the creation of an additional list of harvesting technologies, but the definition of a conceptual structure for explaining why existing technologies are difficult to compare and how they can be positioned using common criteria.

Rather than relying on experimental data or claiming an exhaustive systematic review, the taxonomy is grounded in physical and functional principles reported in the agricultural engineering and harvesting robotics literature. Future work can extend the framework by applying it to larger technology datasets, formalizing quantitative selectivity indicators, and evaluating whether the proposed axes and selectivity attribute improve comparison across crops, production systems, and robotic architectures.

\end{document}